\documentclass[sigconf]{acmart}

\AtBeginDocument{%
  }

\copyrightyear{2024}
\acmYear{2024}
\acmDOI{XXXXXXX.XXXXXXX}

\usepackage[capitalize]{cleveref}
\usepackage{caption}
\usepackage{subcaption}
\usepackage{float}
\usepackage[htt]{hyphenat} 
\usepackage{array}
\usepackage{arydshln}

\usepackage{svg}
\usepackage{algorithm}
\usepackage{algpseudocode}
\usepackage{pifont}
\usepackage{lipsum}  
\usepackage{diagbox}
\usepackage{multirow}
\usepackage[export]{adjustbox}

\crefname{section}{Sec.}{Secs.}
\Crefname{section}{Section}{Sections}
\Crefname{table}{Table}{Tables}
\crefname{table}{Tab.}{Tabs.}

\usepackage{hyperref}
\hypersetup{
    colorlinks=true,
    linkcolor=blue,
    filecolor=magenta,      
    urlcolor=cyan,
    pdftitle={Overleaf Example},
    pdfpagemode=FullScreen,
    }
\usepackage{graphicx}

\acmConference[arXiv]{}{2024}{}

\begin{document}

\title{Low-Latency Scalable Streaming for Event-Based Vision}

\author{Andrew Hamara}
\email{andrew_hamara1@baylor.edu}
\affiliation{%
  \institution{Baylor University}
  \city{Waco}
  \state{Texas}
  \country{USA}
}

\author{Benjamin Kilpatrick}
\email{benjamin_kilpatrick1@baylor.edu}
\affiliation{%
  \institution{Baylor University}
  \city{Waco}
  \state{Texas}
  \country{USA}
}

\author{Alex Baratta}
\email{alex_baratta1@baylor.edu}
\affiliation{%
  \institution{Baylor University}
  \city{Waco}
  \state{Texas}
  \country{USA}
}

\author{Brendon Kofink}
\email{brendon_kofink2@baylor.edu}
\affiliation{%
  \institution{Baylor University}
  \city{Waco}
  \state{Texas}
  \country{USA}
}

\author{Andrew C. Freeman}
\email{andrew_freeman@baylor.edu}
\orcid{0000-0002-7927-8245}
\affiliation{%
  \institution{Baylor University}
  \city{Waco}
  \state{Texas}
  \country{USA}
}

\renewcommand{\shortauthors}{Hamara et al.}

\begin{abstract}
Recently, we have witnessed the rise of novel ``event-based'' camera sensors for high-speed, low-power video capture. Rather than recording discrete image frames, these sensors output asynchronous ``event'' tuples with microsecond precision, only when the brightness change of a given pixel exceeds a certain threshold. Although these sensors have enabled compelling new computer vision applications, these applications often require expensive, power-hungry GPU systems, rendering them incompatible for deployment on the low-power devices for which event cameras are optimized. Whereas receiver-driven rate adaptation is a crucial feature of modern video streaming solutions, this topic is underexplored in the realm of event-based vision systems. On a real-world event camera dataset, we first demonstrate that a state-of-the-art object detection application is resilient to dramatic data loss, and that this loss may be weighted towards the end of each temporal window. We then propose a scalable streaming method for event-based data based on Media Over QUIC, prioritizing object detection performance and low latency. The application server can receive complementary event data across several streams simultaneously, and drop streams as needed to maintain a certain latency.  With a latency target of 5 ms for end-to-end transmission across a small network, we observe an average reduction in detection mAP as low as 0.36. With a more relaxed latency target of 50 ms, we observe an average mAP reduction as low as 0.19.
 
\end{abstract}



\received{22 November 2024}

\maketitle


\section{Introduction}

Modern video streaming systems would be impractical without rate adaptation strategies. These mechanisms can accommodate dramatic changes in network conditions, allowing the receiver to select the stream with the highest data rate it can receive without incurring too much latency. Existing streaming systems, while effective for general use, have the following drawbacks in the context of machine vision. Firstly, the adaptation mechanisms focus on maximizing the user quality of experience (QoE), rather than the performance of a computer vision application. Secondly, they are closely tied to classical video codecs, limiting their use to classical frame-based cameras.

Meanwhile, the computer vision literature is marked by the recent rise of neuromorphic ``event cameras.'' These sensors capture video in a manner completely different than classical sensors. Rather than recording discrete image frames through the synchronous capture of all pixels, these sensors instead record asynchronous change events.  That is, when a given pixel gets brighter or darker by a given amount, it will produce a discrete piece of data with a high-precision timestamp. These sensors have found broad applicability for computer vision applications and robotics systems, owing to their extremely high speed and high dynamic range \cite{gallego_event-based_2022}.

\begin{figure*}
  \centering
    \begin{subfigure}{0.33\textwidth}
        \centering
        \includegraphics[width=\linewidth]{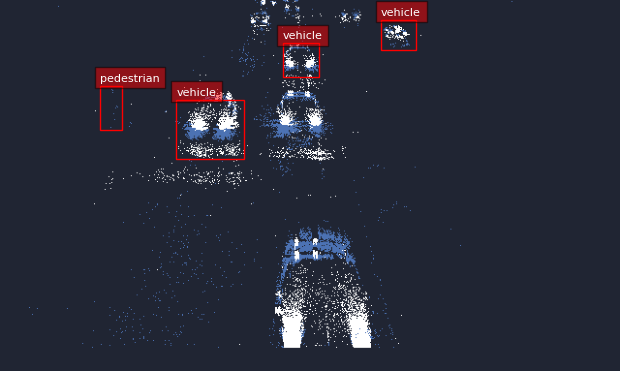}
        \caption{100 Mbps}
        \label{fig:teaser_100}
    \end{subfigure}
    \begin{subfigure}{0.33\textwidth}
        \centering
        \includegraphics[width=\linewidth]{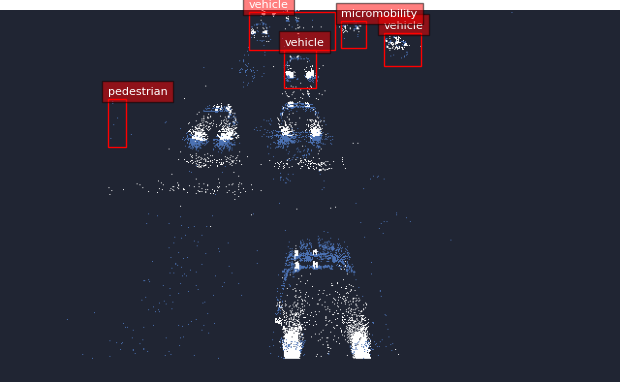}
        \caption{50 Mbps}
    \end{subfigure}
    \begin{subfigure}{0.33\textwidth}
        \centering
        \includegraphics[width=\linewidth]{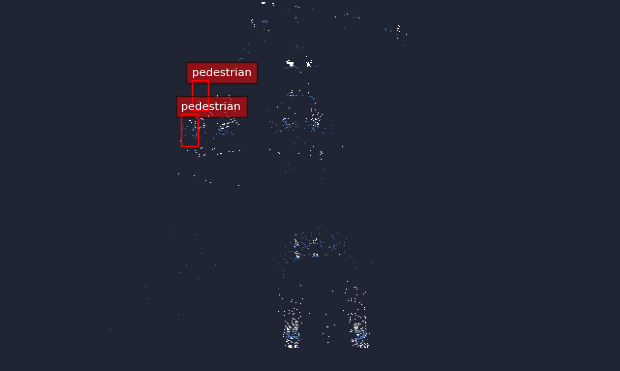}
        \caption{5 Mbps}
    \end{subfigure}
    \caption{Examples of event reduction at various bandwidth limits, with object detections overlaid. At lower bandwidths, there are fewer events in the raw representation, lowering the application accuracy.}
    \label{fig:five_images}
\end{figure*}

To date, no work has thoroughly explored lossy, adaptive streaming of the data from these cameras. This paper makes inroads into this problem, providing the first evaluation of receiver-driven rate adaptation of raw event camera data over a traditional media protocol. We summarize our contributions as follows:

\begin{itemize}
    \item Evaluate the effect of event data loss on object detection accuracy, finding a reduction in mean average precision (mAP) of only 0.17 with 64.9\% data loss.
    \item Evaluate two strategies for partitioning event camera data across multiple streams for scalable coding.
    \item Propose a scalable streaming solution with evenly-sized streams, an MoQ backbone, and a latency-driven stream selection algorithm.
    \item Evaluate our solution with various configurations for stream numbers and network bandwidth. Under a strict latency requirement and at sufficiently high bandwidth levels, we can maintain an average processing latency of less than 4 ms and reduction in mAP of 0.36. Under a less restrictive latency target, we can maintain latency of less than 33 ms with an average mAP reduction of 0.19. 
\end{itemize}


\section{Related work}

\subsection{Event Cameras}

By far, the most common event sensor is the Dynamic Vision System (DVS)\footnote{For the remainder of this paper, we will use ``event sensor'' and ``DVS'' interchangeably.}. A DVS pixel continuously monitors the incident log intensity, firing an event when the change in log intensity exceeds a given threshold \cite{lichtsteiner_128x128_2008}. The output event is a 4-tuple of the form $\langle x, y, t, p \rangle$: the pixel coordinates are conveyed by $x$ and $y$; the microsecond timestamp is conveyed by $t$; and the polarity, $p$, is a single bit denoting the sign of the intensity change \cite{gallego_event-based_2022}. The effective dynamic range of these sensors exceeds 120 db \cite{gallego_event-based_2022}.

Since these sensors capture information only where the intensity is \textit{changing}, they tend to produce events primarily at the moving edges. When the camera is motionless in the absence of noise, only the edges of moving objects will induce events. When the camera is moving, however, many pixels will fire in rapid succession, and it is difficult to separate the camera's ego-motion from that of objects in the scene \cite{mueggler_event-camera_2017,vidal_ultimate_2018,chamorro_event-based_2022,jiao_comparing_2021}.

\subsection{Event-Based Vision}

Event cameras have been leveraged for many vision applications, including object segmentation \cite{biswas_halsie_2024,chen_segment_2023, Stoffregen_2019_ICCV, stoffregen_2018}, gesture recognition \cite{liu_fast_2022,gao_action_2023,gao_hypergraph-based_2024}, and object detection \cite{cannici_asynchronous_2019,gehrig_recurrent_2023,ramesh_pca-rect_2019, deng_2020, pradhan_2019}. A major benefit to event cameras is their low power usage, allowing them to be deployed on lightweight robotic vehicles \cite{nair_enhancing_2024,vitale_event-driven_2021,dimitrova_towards_2020}. However, the computational complexity of the applications often obviates this benefit. Although some researchers have implemented applications in custom neuromorphic hardware \cite{vitale_event-driven_2021}, most vision-oriented work either processes data offline \cite{mueggler_2018, ramesh_2020, milde_2015, censi_2014} or leverages power-hungry GPUs \cite{Zhu_2018, Bardow_2016_CVPR, Zhu_2019_CVPR}. When GPUs are necessary, it can become impractical to have this hardware onboard a robotic vehicle.

A growing number of researchers focus on forward-looking applications with spiking neural networks (SNNs) \cite{vedaldi_learning_2020, Yao_2021_ICCV, Zhang_2022_CVPR, cordone_spiking}. These networks more closely emulate biological information processing, where neurons communicate through discrete spikes. Similar to RNNs, SNN neurons maintain a temporal state, but only generate spikes at certain thresholds. The non-differentiability, and thus incompatibility with backpropagation \cite{backprop}, of the spiking mechanism has introduced optimization challenges in SNN applications \cite{bellec2020solution, lee2020enabling, gradient_snn, slayer}. Overall, SNNs require further development to overcome optimization challenges and demonstrate competitive performance in real-world applications.

As such, classical deep learning has driven significant advances in object tracking for traditional frame-based cameras \cite{siamese, Nam_2016_CVPR, Li_2019_ICCV, Gao_2020_CVPR, Danelljan_2017_CVPR, Bhat_2019_ICCV}, inspiring event-based methods that group events into temporal "frames" for compatibility with convolutional and attention-based computations \cite{gallego_event-based_2022, Tulyakov_2019_ICCV, li_j, barchid, baldwin, Maqueda_2018_CVPR, rebecq_high_2021}. Optimizations of self-attention mechanisms in vision transformers for frame-based tasks \cite{maxvit} have improved computational efficiency and enabled event-based researchers to achieve state-of-the-art runtime and precision in object detection, surpassing convolutional counterparts \cite{gehrig_recurrent_2023}. Early works treated frames independently for object detection, often discarding temporal information crucial for slower-moving objects \cite{iacano_IROS, jiang_ICRA, Chen_2018_CVPR_Workshops}. Recent advancements address this limitation by incorporating recurrent neural networks (RNNs) to capture and retain long-term temporal dependencies \cite{NEURIPS2020_perot, li_j, gehrig_recurrent_2023}.

\begin{figure*}
    \centering
    \includegraphics[width=1\linewidth]{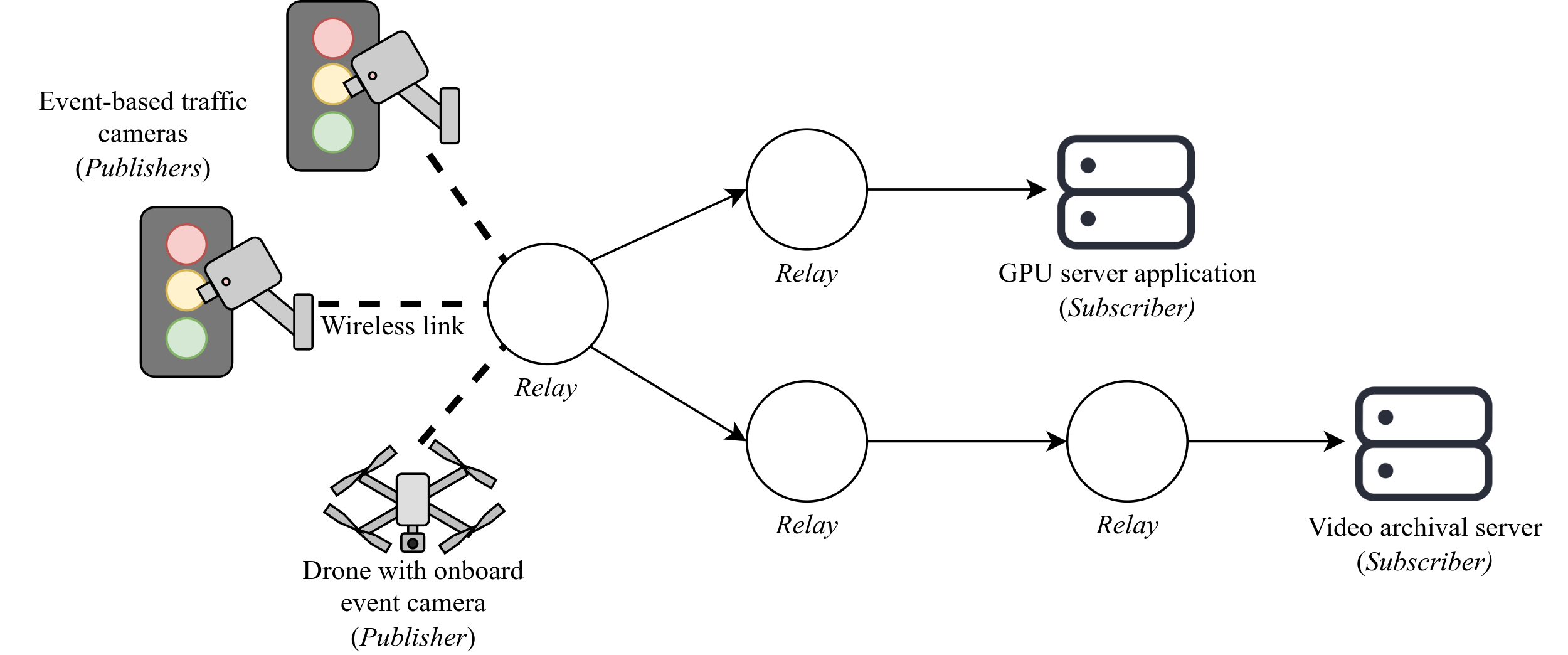}
    \caption{Example of a ``smart city'' video system made possible by scalable streaming with MoQ. Heavy vision application computation can be offloaded from heterogeneous edge sensors, prioritizing low latency and receiving a subset of the published data. Meanwhile, a video archival server can receive all the data published by the cameras, since it is more tolerant to spikes in latency.}
    \label{fig:system_overview}
\end{figure*}

\subsection{Event Compression and Streaming}\label{sec:related_event_compression_streaming}

Even at low resolutions, event sensors can produce millions of events per second. If one wants to send the data offboard for application processing, some form of compression is necessary. Khan et al. proposed a method for ``lossless'' DVS compression which used polarity event frames (counting the number of events of each polarity, for each pixel) in conjunction with a frame-based video encoder \cite{khan}. This method is actually \textit{lossy} in the generation of the event frames, due to temporal quantization. Additionally, it has poor performance at high frame rates, due to the overhead of classical high-frame-rate encoding. Similar lossy compression methods for DVS either quantize the temporal components \cite{towards_dvs_lossy,Schiopu_1} or discard events entirely.  Recent work has explored lossless DVS compression designed around on-camera hardware implementations \cite{Schiopu_2,Schiopu_3}, point cloud geometry \cite{martini_lossless_2022}, and a learning-based entropy model \cite{sezavar_learning-based_2024}. 

Fischer et al. explored the effect of spatially subsampling DVS events for a place recognition application, demonstrating competitive performance with a small subset of the original data \cite{fischer_how_2022}. In a similar vein, Glover et al. proposed a framework for adjusting the application processing rate according to the received event rate \cite{glover_event-driven_2018}. On the other hand, Delbruck et al. proposed a feedback mechanism based on camera intrinsics for adjusting the event rate produced by the camera \cite{delbruck_feedback_2021}. These works highlight the highly variable nature of event sensing: more motion in the video yields more events, and more events will slow the transmission and processing speed. 

The work most closely related to this paper is that of Banerjee et al. The authors proposed a lossy compression system \cite{banerjee_lossy_2020} and data rate optimization scheme for an online object tracking application \cite{banerjee_joint_2024}. However, their method requires the complementary intensity information from a classical framed sensor, it requires a strict integration of the particular application, and it is not amenable to streaming across multiple network nodes or to multiple receivers.

Overall, compression and streaming are severely understudied topics in the event camera literature. Much research focuses primarily on developing accurate applications, regardless of the data requirements or speed.

\subsection{Classical Video Streaming}

In contrast, the issues of compression and streaming are well-understood and vital components in classical video systems.

Early efforts in rate-adaptive streaming included scalable video coding (SVC). Here, one can encode a source video with enhancement layers to support variable-rate streaming \cite{mccanne_scalable_nodate,schwarz_overview_2007}. A client ingests a base layer for a low-quality stream and subsequent layers increase the visual quality, temporal resolution, or spatial resolution. Higher layers are additive and dependent on lower layers. When bandwidth is limited, the client simply receives fewer enhancement layers, reducing the bitrate and quality.

SVC has seen some success, most notably in its incorporation into H.264 and WebRTC. In recent years, however, adaptive bitrate (ABR) streaming has proven to be the dominant market choice for variable-rate streaming, finding widespread support in HLS and DASH. In the ABR paradigm, the server encodes the source video at various quality levels, producing distinct, independent streams. The client receives only one stream at a time, but can switch to a different stream if network conditions change.

Media over QUIC (MoQ) is a work-in-progress protocol aimed at simplifying low-latency video streaming \cite{gurel_media-over-quic_2024}. It allows video streams to be published across multiple ``tracks,'' and a receiver may select between these tracks according to its needs and network conditions. MoQ supports multi-receiver broadcast and adaptation through the use of a relay system. Receivers simply connect to a relay and subscribe to the desired track(s). Importantly for our purposes, MoQ provides straightforward mechanisms to send generic data (not only classical video), so long as it is time-ordered.

\section{Problem Statement}

As noted above, much of the existing event vision literature aims to develop low-latency applications with bespoke, onboard hardware implementations. The other major vein of research focuses on high-accuracy applications, sacrificing speed and power requirements by using expensive, high-power GPU devices. In the latter cases, a fundamental question remains largely ignored: \textit{how does the data get from the event camera to the server?} 

We envision a scenario where these cameras may be deployed in the real world for ``smart city'' traffic pattern analysis and drone surveillance, as illustrated in \cref{fig:system_overview}. A small drone vehicle cannot meet the power and weight requirements for a GPU system. Likewise, edge GPU resources on traffic cameras will be underutilized during periods of low motion. For practicality, cost, maintenance, and improvement purposes, we instead want to process the camera data on a remote server in a distant, secure location. It must receive some subset of the event video with low latency, so that it can quickly compute application results for mission-oriented response. Alongside the low-latency application, we may want another server to archive \textit{all} of the data produced by the camera, regardless of any latency incurred, for post-facto applications and monitoring. While we could convert the events at the sensors into a low-rate framed representation, this is not amenable to applications that aim for high-rate performance.


The key problem is to design a generic, multi-consumer system that can handle different performance requirements for both low-latency and latency-tolerant applications. Key technical challenges include:

\begin{itemize}
\item Efficiently partitioning the source event stream to serve the needs of latency-intolerant consumers.
\item Dynamically responding to changes in compute, bandwidth, and source video data rates. 
\item Ensuring robust, fault-tolerant delivery.
\end{itemize}

Solving this problem would enable a versatile computer vision system that can simultaneously support real-time use cases like obstacle avoidance or navigation for robotic vehicles, as well as retrospective video analysis and archiving. Additionally, a solution would provide much more utility to existing lossless event compression algorithms, which could be applied only to small partitions of the event stream, rather than the entire event stream at once.

\section{Event Vision Performance Under Loss}\label{sec:fixed_bitrate}

As our goal was to develop strategies for partitioning an event camera stream into various rate levels, we first needed to evaluate the effect of data loss (discarding events) on application performance. For this work, we selected the state-of-the-art Recurrent Vision Transformer (RVT) object detection model \cite{gehrig_recurrent_2023}. This model achieves high mean average precision (mAP, at 50+\% intersection over union of the bounding boxes) with recurrence layers to maintain a feature state between time windows \cite{gehrig_recurrent_2023}. Furthermore, it achieves this high accuracy with $< 12$ ms inference time per time window on an NVIDIA T4 GPU \cite{gehrig_recurrent_2023}.

Specifically, we opted to use an updated version of RVT that was trained on a traffic monitoring dataset, eTraM \cite{verma_etram_2024}. The eTraM dataset contains 10 hours of footage from a stationary event camera pointed at intersections and roads \cite{verma_etram_2024}. The videos were captured during daylight and nighttime settings at a resolution of 1280×720 pixels, and the object classes include vehicles, pedestrians, and micro-mobility devices (bicycles, wheelchairs, and scooters). We selected a random subset of the test dataset (not used for training), consisting of 5 daytime scenes and 5 nighttime scenes, each approximately 3 minutes in duration. Of this subset, the raw (uncompressed) file size ranges from 459.6 MB to 10.7 GB. This wide range reveals a fundamental difference from standard video, where the size of each uncompressed image is constant. As described in \cref{sec:related_event_compression_streaming}, a higher level of motion in the video produces more events and a larger raw representation.

First, we wished to evaluate the effect of data loss on object detection performance. One cannot simply discard some fixed percentage of events, because the event rate produced by the camera is wholly dependent on the level of motion in the scene. Therefore, we implemented an event reduction scheme with a fixed bandwidth parameter. The program groups the events into 50 ms time windows (as ingested by RVT), and simply discards any events that would exceed the bandwidth threshold for each window. We then performed inference on the resulting files with RVT, using the pre-trained model from the authors of eTraM \cite{verma_etram_2024}.

\begin{figure}
    \centering
    \includegraphics[width=1\linewidth]{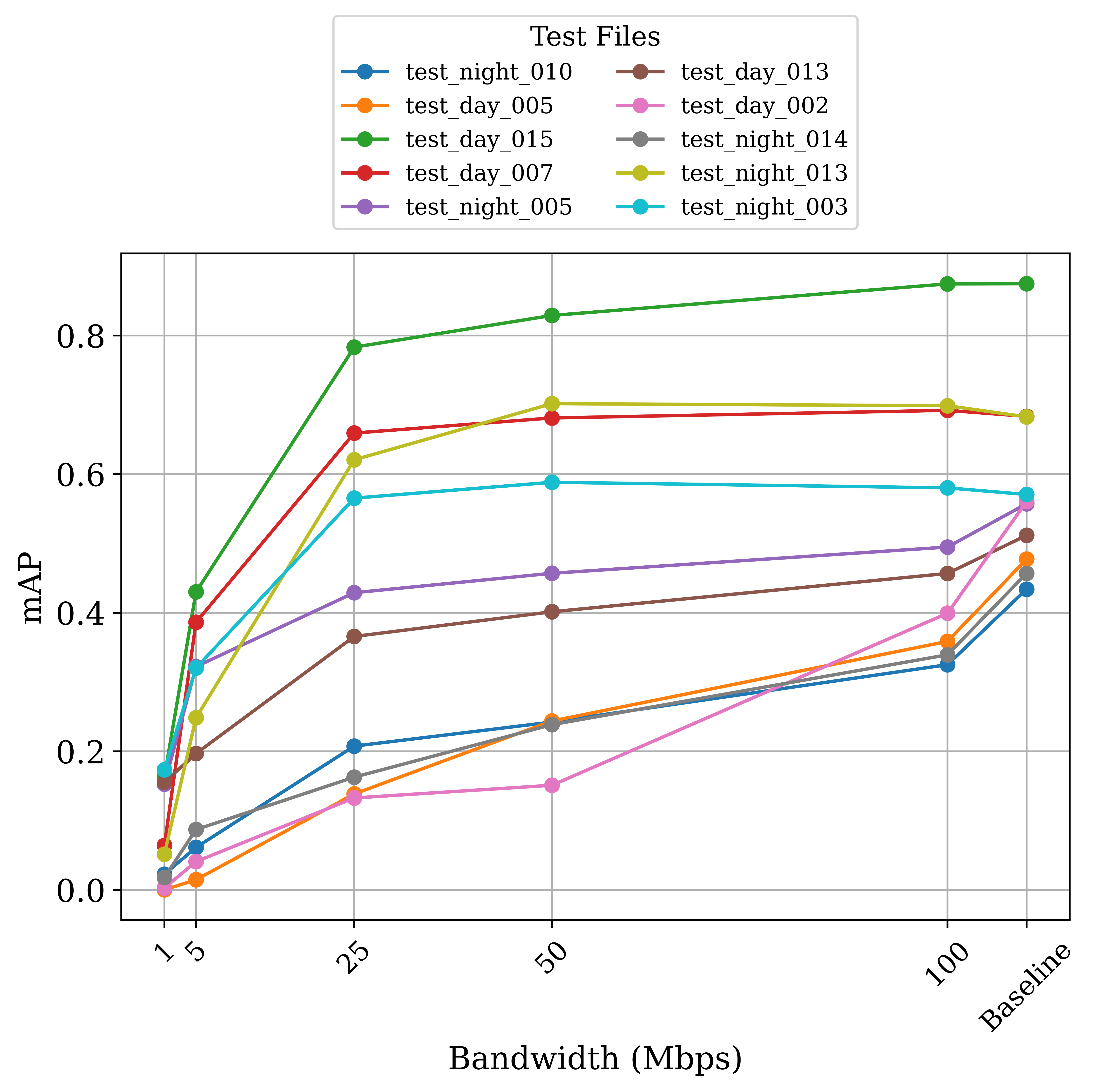}
    \caption{The change in object detection accuracy at various bandwidth levels for each video in our test dataset, without latency-driven rate adaptation (\cref{sec:fixed_bitrate}). Object detection performance remains high even when the majority of the events are removed. }
    \label{fig:fixed_bitrate_map}
\end{figure}

We found that the RVT model is rather resilient to data loss, as shown in \cref{fig:fixed_bitrate_map,tab:fixed_bitrate}. Specifically, we observe that the average mAP is reduced by only 0.17 with a 64.9\% loss in data. At lower data rates, the mAP drops more dramatically, lowering by 0.50 under the 1 Mbps scenario. We offer qualitative results in \cref{fig:five_images}. Here, one can clearly see the degradation in object detection accuracy as the bandwidth decreases.

\begin{table}
    \centering
    \begin{tabular}{p{0.15\linewidth}||p{0.1\linewidth}p{0.1\linewidth}p{0.1\linewidth}|p{0.1\linewidth}p{0.1\linewidth}}
        Bandwidth   & \multicolumn{3}{c}{Latency (ms)} & Mean data & Mean mAP \\ 
        (Mbps)  & Max & Median & Mean & loss (\%) & difference \\ \hline
         1   &  537.2  & 33.9 & 37.5 & 98.1 & 0.50 \\
         5   &  892.5  & 17.3 & 73.8 & 91.1 & 0.37 \\
         25  &  989.3  & 37.9 & 55.4 & 64.9 & 0.17 \\
         50  &  1241.2 & 44.3 & 35.9 & 40.9 & 0.13 \\
         100 &  1330.8 & 34.9 & 34.6 & 7.9 & 0.06 \\
    \end{tabular}
    \caption{Summary of the experimental results from the fixed-rate event loss. Without rate adaptation, even an extremely lossy transmission incurs latency that is too high for real-time event vision systems.}
    \label{tab:fixed_bitrate}
\end{table}

\begin{figure*}
    \centering
    \begin{subfigure}{0.49\textwidth}
        \centering
        \adjincludegraphics[height=10cm,trim={0 0 {.5\width} 0},clip]{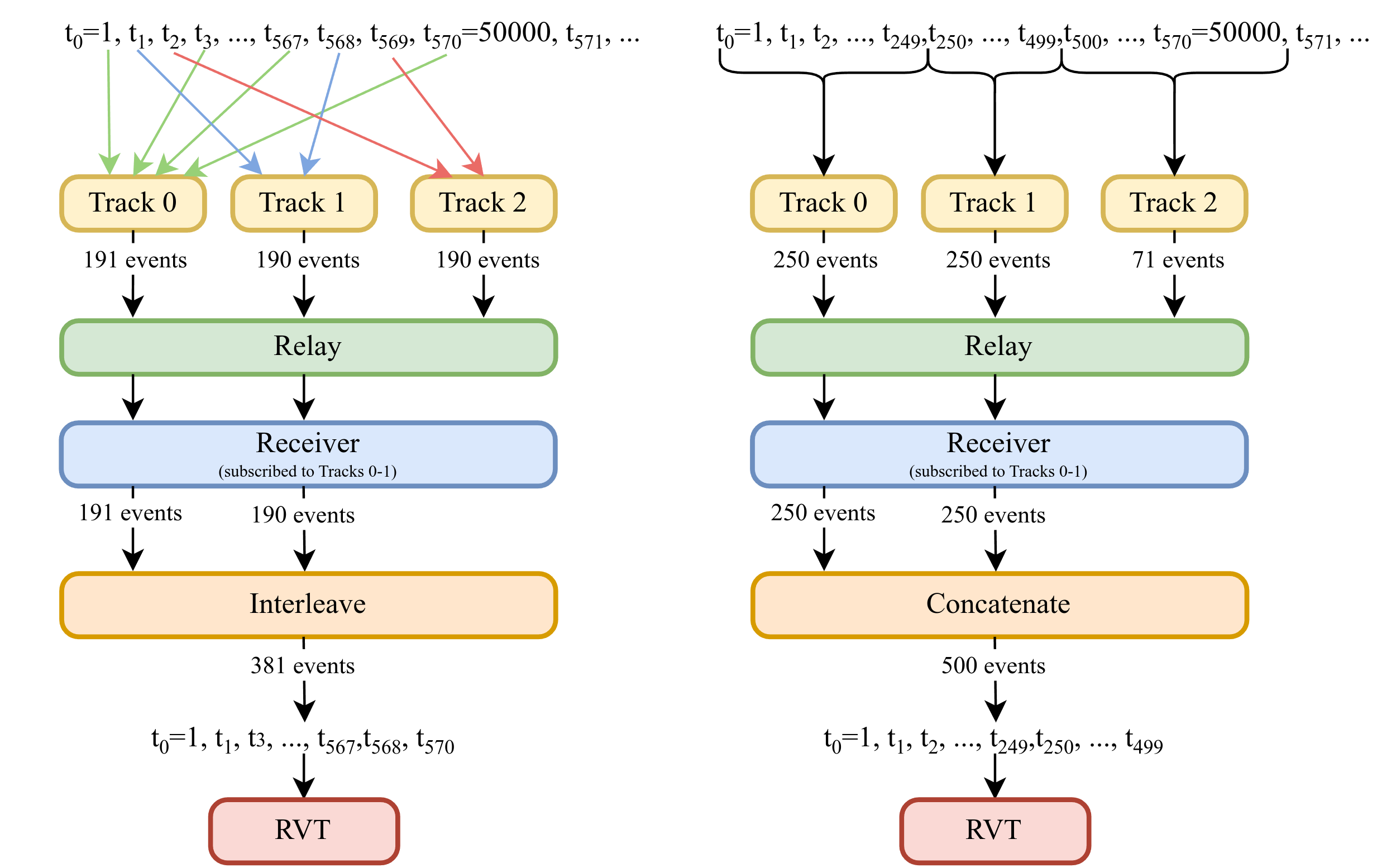}
        \caption{Even event partitioning}
    \end{subfigure}
    \begin{subfigure}{0.49\textwidth}
        \centering
        \adjincludegraphics[height=10cm,trim={{.5\width} 0 0 0},clip]{images/event_track_division_both.png}
        \caption{Sequential event partitioning, with $E=250$}
    \end{subfigure}
    \caption{Example of event partitioning and reconstruction with the two strategies described in \cref{sec:event_division}. The receiver is subscribed to only 2 tracks, so the application receives a subset of the source data for inference. The (a) strategy provides a more even distribution of events over the time window, at the cost of enormous latency to interleave the streams for reconstruction. Meanwhile, the (b) strategy has little overhead for reconstruction and the application performance is only negligibly worse.}
    \label{fig:event_track_division}
\end{figure*}

While RVT operates on 50 ms time windows, the representation used internally is a 3D tensor of shape $(2T, H, W)$, where there are $T=10$ further temporal subdivisions (into 5 ms windows) \cite{gehrig_recurrent_2023}. That is, each element of the tensor is the sum of all the events of a given polarity that occurred at the given pixel coordinates across a 5 ms window. Our above experiment merely discarded any events exceeding the given bandwidth threshold, meaning that any loss is weighted towards the end of each 50 ms window. With this in mind, we ran a secondary experiment with event loss evenly distributed across the time window. At bandwidth levels 25 Mbps and 50 Mbps, we saw a relative increase in mAP of only 0.008 and 0.024, respectively. Therefore, the temporal distribution of events within each 50 ms window is not a major contributor to the level of detection accuracy.

\section{Scalable Event Streaming}

Although scalable video coding showed promise in classical multimedia systems, its practicality was hampered by the additional encoding overhead induced by each enhancement layer. In DVS event video, however, we are not tied to a synchronous representation for encoding. Therefore, we can simply partition the data into separate data streams for encoding and transmission. The receiver then decodes each stream and merges them for application inference. The merge process simply requires that each event is placed in the correct temporal order, such that the timestamps of the event sequence increase monotonically.

\subsection{Event Partitioning}\label{sec:event_division}

We explored two strategies to partition the event streams into $N$ different tracks for streaming with MoQ.

For our first strategy, we evenly distribute the camera event stream among the tracks. That is, the $n$th event produced by the camera will be sent across track $(n-1) \bmod N$. To perform application inference, the receiver must reconstruct a single event stream by interleaving the received events. Even though the event order is known, this interleaving operation adds significant overhead at the high data rates of DVS video. We saw up to a 26-second increase in average latency from this event partitioning strategy, making it a poor choice for our low-latency goals. By avoiding this method, we sacrifice only a marginal amount of object detection performance, as described in \cref{sec:fixed_bitrate}.

Instead, we treat the tracks as a bucket system. Each track can receive a fixed number of events, $E$, for any time window. The publisher simply adds the first $E$ events for the time window to Track 0, the second $E$ events to track 1, and so on. This process may leave one track partially unfilled, and some tracks may have no events at all. Additionally, for any given time window, the total number of events published cannot exceed $NE$. However, the stream reconstruction process becomes trivially fast, as the decoded event sequences can be simply copied to consecutive memory locations. We illustrate both stream partitioning and reconstruction methods in \cref{fig:event_track_division}.

\begin{table*}
    \centering
    \begin{tabular}
    {p{0.14\linewidth}p{0.10\linewidth}||p{0.10\linewidth}p{0.10\linewidth}p{0.10\linewidth}|p{0.15\linewidth}p{0.15\linewidth}}
    & & \multicolumn{3}{c}{Latency (ms)} & \multicolumn{1}{l}{Mean through- } & \multicolumn{1}{l}{Mean mAP} \\
         Bandwidth (Mbps) & No. tracks & Max & Median & Mean & put (Mbps) & difference\\ \hline
         \multirow{3}{*}{1} & 5  & \textbf{148.9}  & 41.7 & 37.7       & \textbf{0.79}     & 0.53   \\
                            & 10 & 188.1           & 24.1 & \textbf{27.7}      & 0.69      & \textbf{0.52}   \\
                            & 25 & 87725.0         & 25623.6 & 30493.3 & 0.66              & 0.57   \\ \hline
         \multirow{3}{*}{5} & 5 & \textbf{156.9}   & 9.8 & \textbf{22.8}   & \textbf{0.91} & 0.54   \\
                             & 10 & 506.2          & 20.2 & 45.3     & 0.80                & \textbf{0.52}   \\
                             & 25 & 31549.7        & 4397.9 & 7072.1 & 0.67                & 0.57   \\  \hline
         \multirow{3}{*}{25} & 5 & \textbf{141.2}  & 4.4 & 6.6       & \textbf{3.09}       & \textbf{0.41}   \\
                             & 10 & 189.4          & 9.0 & 8.0      & 1.21                 & 0.51   \\
                             & 25 & 230.2          & 0.6 & \textbf{6.0}      & 1.65        & 0.51   \\ \hline
        \multirow{3}{*}{50} & 5 &   108.2          & 2.4 & 2.8      & \textbf{3.18}        & \textbf{0.41}   \\
                             & 10 & \textbf{105.4} & 5.1 & 5.1      & 1.61                 & 0.49   \\
                             & 25 & 134.7          & 0.4 & \textbf{1.7}      & 2.49        & 0.50   \\ \hline
         \multirow{3}{*}{100} & 5 & \textbf{85.5}  & 1.3 & \textbf{1.3}       & 3.19       & 0.41    \\
                             & 10 & 91.1           & 2.8 & 3.5       & \textbf{6.15}       & \textbf{0.36}    \\
                             & 25 & 121.2          & 0.5 & 1.8      & 3.49                 & 0.48     \\
    \end{tabular}
    \caption{Strict latency results with $E=250$ events per track. Numbers in bold indicate the best performance for each bandwidth level.}
    \label{tab:adapt_results}
\end{table*}

\begin{table*}
    \centering
    \begin{tabular}
    {p{0.14\linewidth}p{0.10\linewidth}||p{0.10\linewidth}p{0.10\linewidth}p{0.10\linewidth}|p{0.15\linewidth}p{0.15\linewidth}}
    & & \multicolumn{3}{c}{Latency (ms)} & \multicolumn{1}{l}{Mean through- } & \multicolumn{1}{l}{Mean mAP} \\
         Bandwidth (Mbps) & No. tracks & Max & Median & Mean & put (Mbps) & difference\\ \hline
         \multirow{1}{*}{25} & 5  & 8349.9  & 56.4 & 173.5       & 11.04 & 0.31   \\ 
        \multirow{1}{*}{50} & 5   &   193.7& 30.0  & 43.2      & 17.14   & 0.24   \\   
         \multirow{1}{*}{100} & 5 & 190.6   & 15.2 & 32.5       & 20.53 &  0.19   \\
    \end{tabular}
    \caption{Relaxed latency results with $E=2500$ events per track.}
    \label{tab:relaxed_results}
\end{table*}

\begin{figure*}
    \centering
    \begin{subfigure}{0.49\linewidth}
        \centering
        \includegraphics[width=\linewidth]{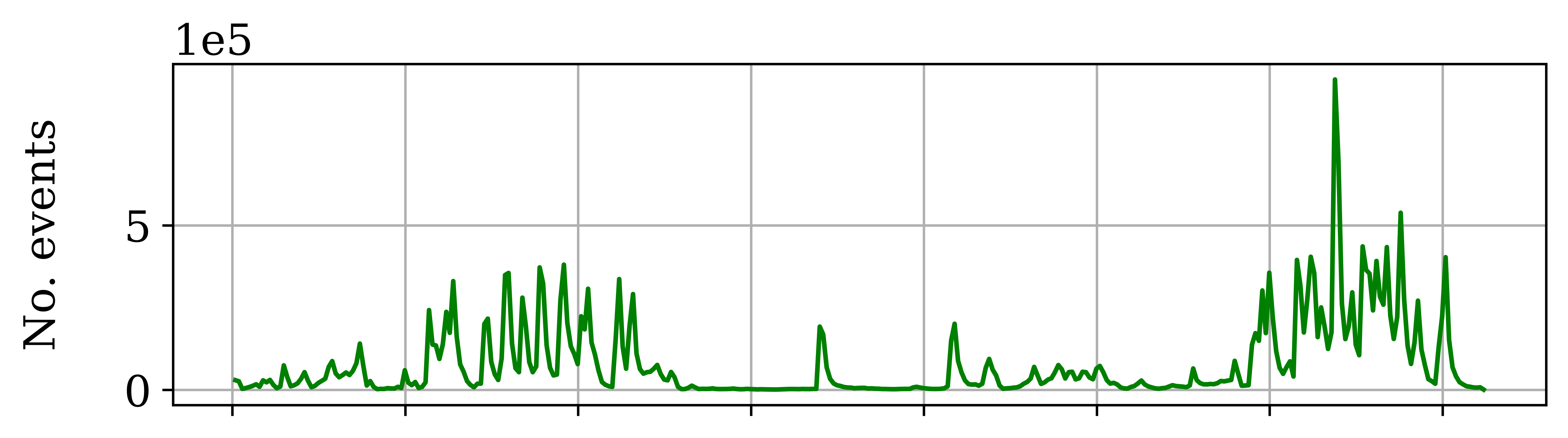}
        \caption{Source video event rate}
        \label{fig:time_example_source_rat}
    \end{subfigure}
    \begin{subfigure}{0.49\linewidth}
        \centering
        \adjincludegraphics[width=\linewidth,trim={0 {0.66\height} 0 0},clip]{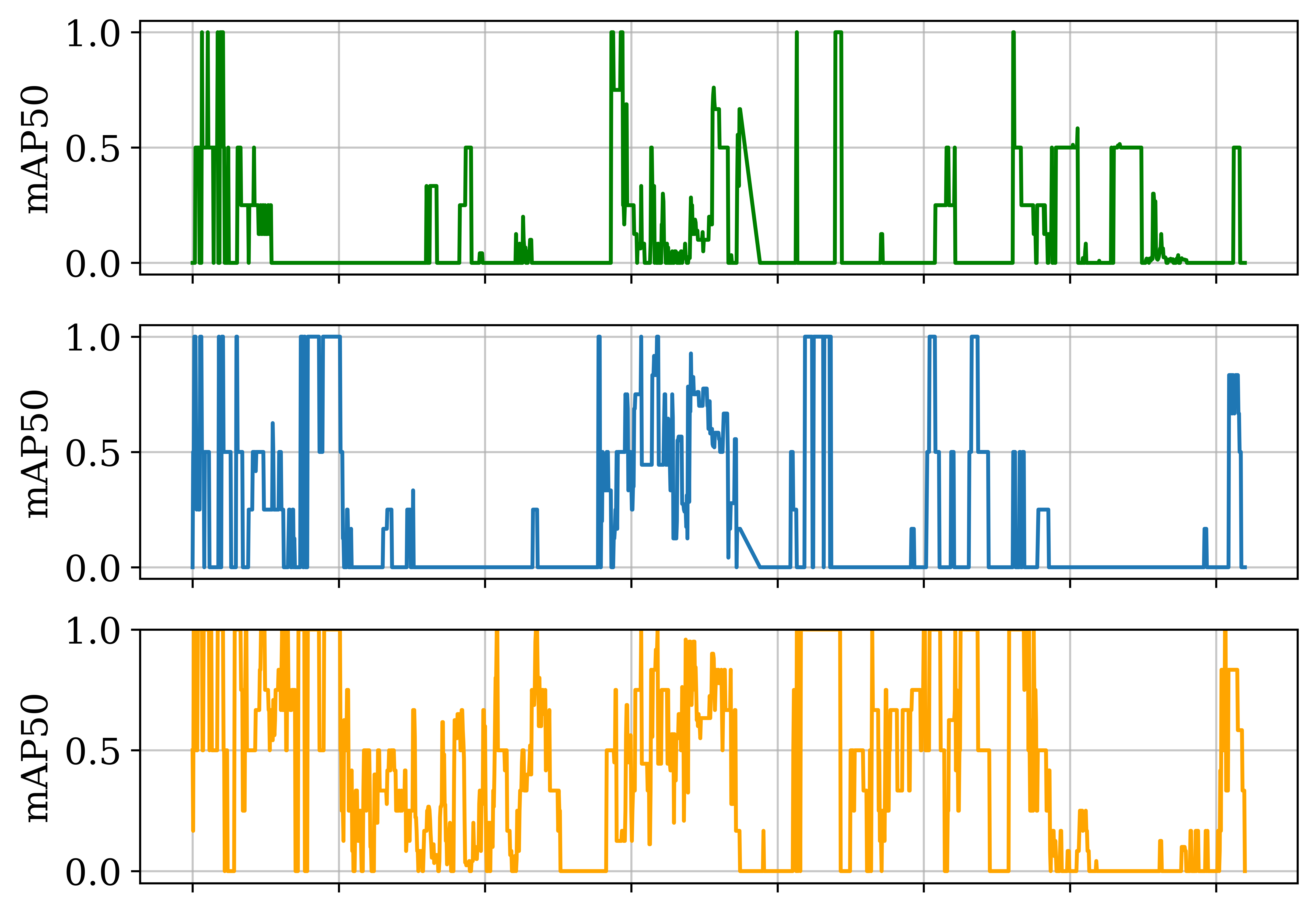}
        \caption{Baseline mAP of source video}
        \label{fig:time_example_source_map}
    \end{subfigure}
    \begin{subfigure}{0.49\linewidth}
        \centering
        \includegraphics[width=\linewidth]{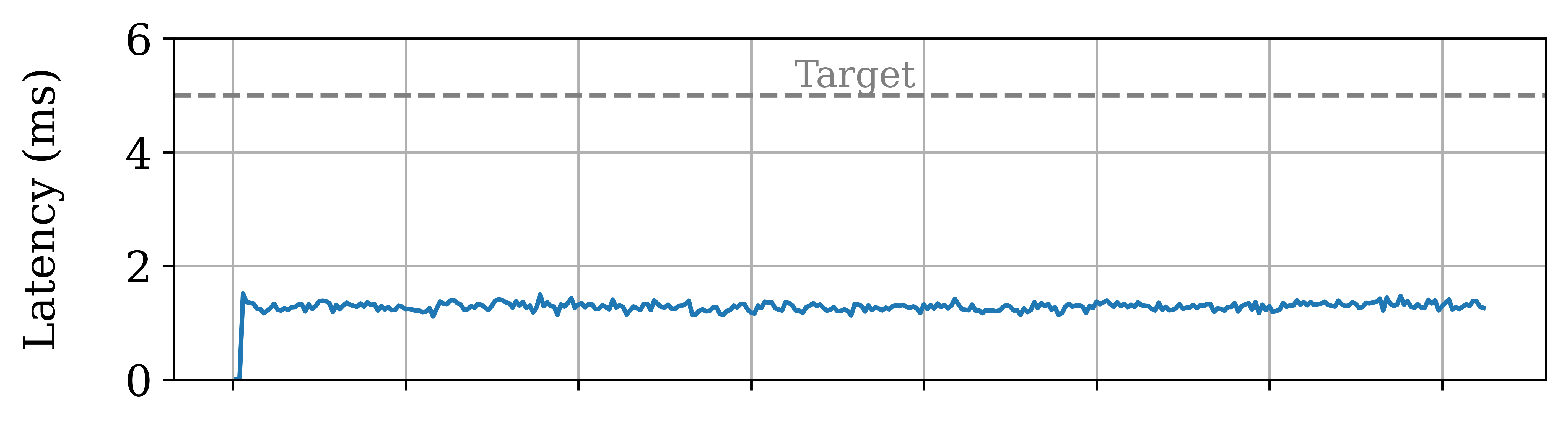}
        \caption{Strict - latency}
    \end{subfigure}
    \begin{subfigure}{0.49\linewidth}
        \centering
        \includegraphics[width=\linewidth]{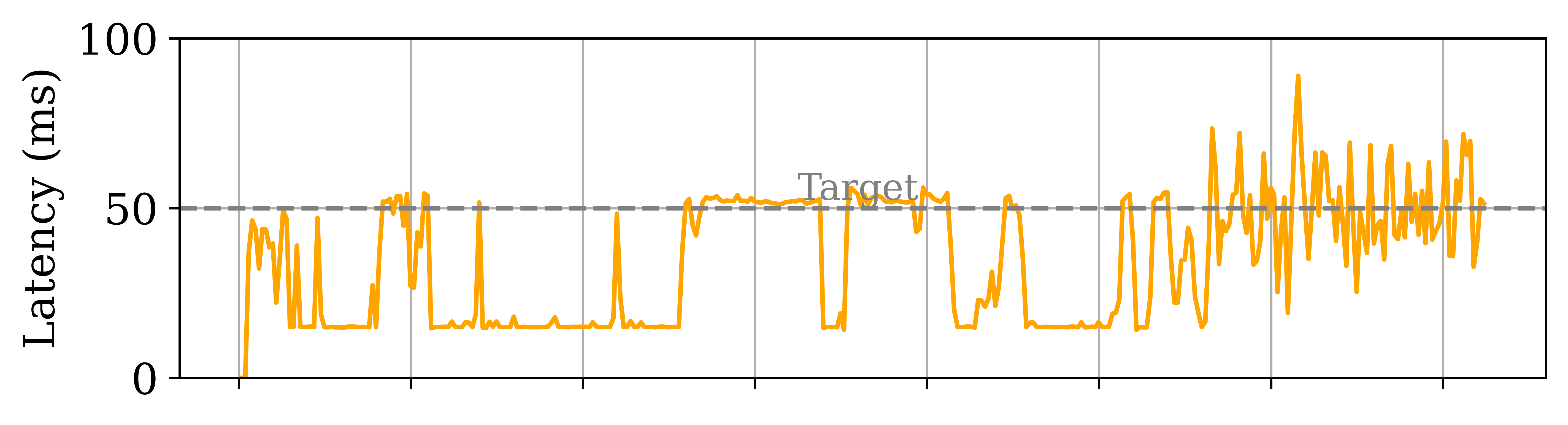}
        \caption{Relaxed - latency}
        \label{fig:time_example_relaxed_latency}
    \end{subfigure}
    \begin{subfigure}{0.49\linewidth}
        \centering
        \includegraphics[width=\linewidth]{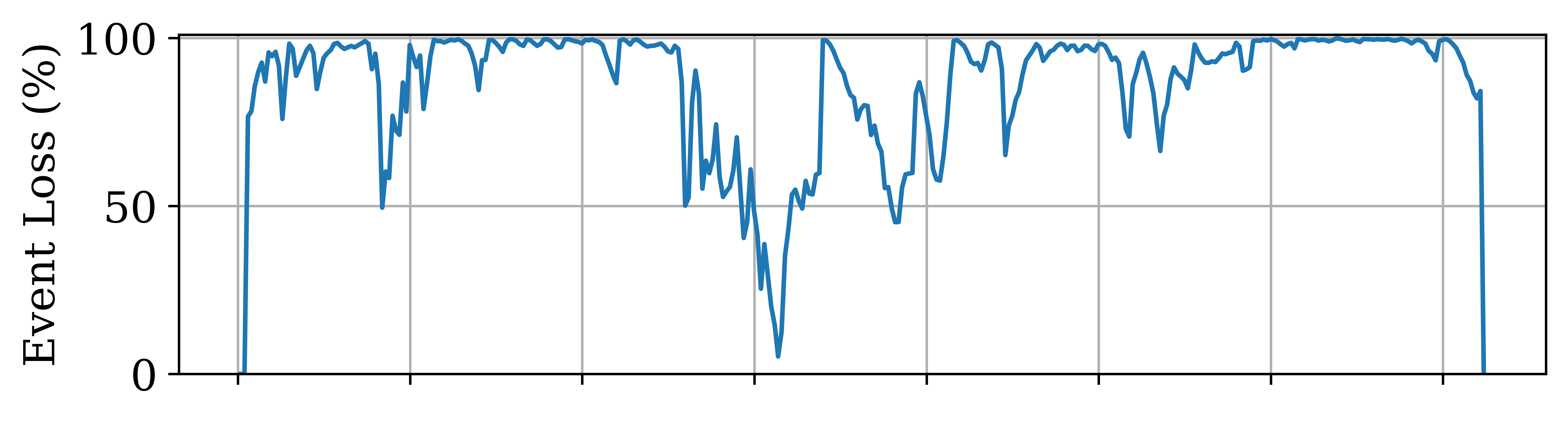}
        \caption{Strict - event loss rate}
        \label{fig:time_example_strict_loss}
    \end{subfigure}
    \begin{subfigure}{0.49\linewidth}
        \centering
        \includegraphics[width=\linewidth]{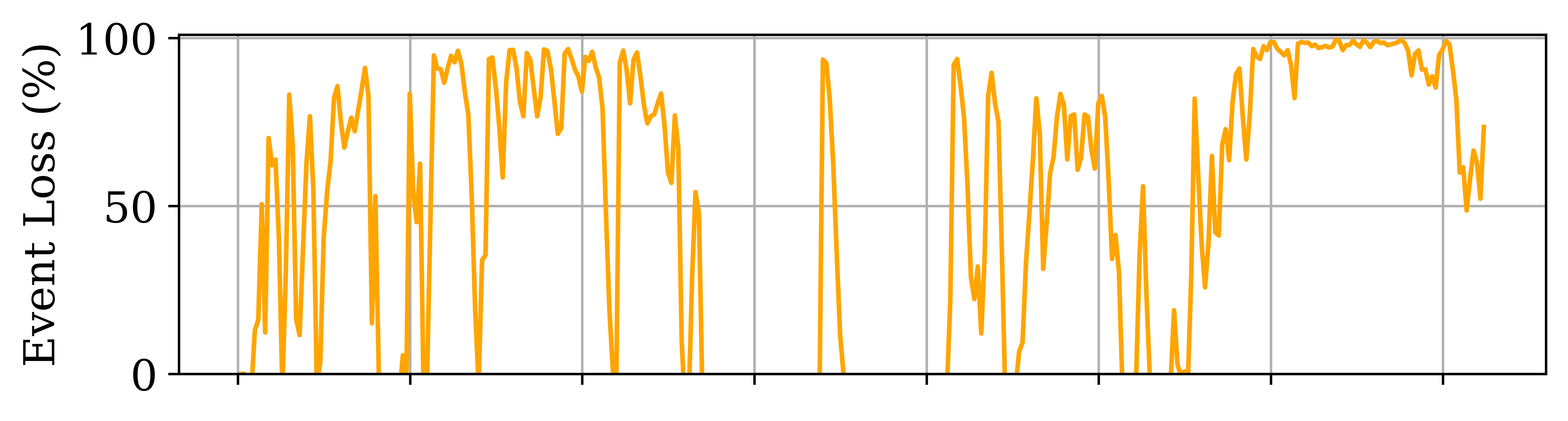}
        \caption{Relaxed - event loss rate}
        \label{fig:time_example_relaxed_loss}
    \end{subfigure}
    \begin{subfigure}{0.49\linewidth}
        \centering
        \adjincludegraphics[width=\linewidth,trim={0 {0.33\height} 0 {0.33\height}},clip]{images/time/13_100_5_map.png}
        \caption{Strict - mAP}
        \label{fig:time_example_strict_map}
    \end{subfigure}
    \begin{subfigure}{0.49\linewidth}
        \centering
        \adjincludegraphics[width=\linewidth,trim={0 0 0 {0.66\height}},clip]{images/time/13_100_5_map.png}
        \caption{Relaxed - mAP}
        \label{fig:time_example_relaxed_map}
    \end{subfigure} 
    \caption{Experimental results from sample video \texttt{test\_day\_013} with $N=5$ tracks and $B=100$ Mbps, showing various metrics across the timespan of the video. These results demonstrate a case where the event loss induced by our streaming system actually \textit{increases} the mAP for many time segments. The increase in the source data rate towards the end of the video corresponds to an increase in latency for the relaxed configuration and a dropoff in mAP for both the strict and relaxed configurations.}
    \label{fig:time_example}
\end{figure*}

\subsection{Track Selection}\label{sec:track_selection}

As with SVC, our tracks are complementary, such that receiving more tracks will yield a higher quality video stream. A crucial part of our system, then, is the track selection process. The receiver should maximize the data received, while staying near or below a certain latency target.

Our track selection strategy is a rate adaptation mechanism designed to dynamically adjust a ``chunk'' size based on measured throughput. Once we receive the next data segment for each subscribed track, we measure the latency as the time elapsed between the first track's send time and the receiver's stream reconstruction. If the measured latency is above our target latency, $L$, we decrease our chunk size by 20\%. In contrast, if the measure latency is less than our target latency, we increase the chunk size by 20\%. Since each track will transmit no more than $E$ events per segment (and only the highest track with data received may transmit less than $E$ events), we may simply divide the resulting chunk size by $E$ to determine the optimal number of track subscriptions. We then subscribe or unsubscribe as needed so that our subscription count matches the target.

There is an inherent delay from MoQ that comes with each new track subscription. To avoid stalling, we only subscribe to one new track at a time. We further wait until the subscription handshake has completed before we may subscribe to another new track. In contrast, we can \textit{un}subscribe from many tracks at once. This functions as a congestion avoidance mechanism at the receiver, with fast recovery when latency is high.

Although traditional SVC/ABR adaptation algorithms give high weight to data rate stability and the human QoE, our algorithm chiefly targets latency optimization. For a latency-oriented vision application, thrashing in the received data rate (and therefore video ``quality'') is not a major concern; the application simply needs as much data as possible within the bounds of the latency requirements. This is especially true for an event-based application, where human viewership is of scant utility.



\subsection{Implementation and Experimental Setup}
We implemented our system in Rust\footnote{Source code will be released upon publication}, using the \texttt{moq-rs} library to provide the MoQ publisher, subscriber, and relay functionality \cite{kixelated_kixelatedmoq-rs_2024}. Since we use a pre-recorded dataset for our experiments, we simulate the timing of a live, connected event camera. That is, we insert a variable time delay between sending each event window, to match the real-world time that the camera produced the events. A unique thread manages each track on the receiver. The main thread performs the latency-based rate adaptation algorithm (\cref{sec:track_selection}), and passes messages to the track threads to subscribe or unsubscribe as needed. The track threads then communicate any data received to the main thread, which reconstructs the stream as described in \cref{sec:event_division}.

For the following experiments, we evaluated our scalable streaming system on the 10-video event dataset used in \cref{sec:fixed_bitrate}. We set up one MoQ relay, running in a Docker container. Using the Linux traffic control (\texttt{tc}) and token bucket filter (\texttt{tbf}) utilities, we set various bandwidth limitations, $B$. For each bandwidth value, we set the matching ``burst'' level to simulate realistic variations in data throughput. As this initial work focuses on the effect of event partitioning (rather than event compression), we transmit each event as raw, uncompressed binary, spanning 16 bytes. We performed object detection inference with RVT using the reference model weights provided by the authors \cite{gehrig_recurrent_2023}. The key metrics we gathered included receiver latency, data throughput, and application accuracy. All experimental runs were performed on an Intel Core i9-14900K with Ubuntu 24.04 and \texttt{moq-rs} version 0.5.2.

Below, we present results from evaluations with end-to-end target latencies of 5 ms and 50 ms across our simulated network. Since a real-world network will likely have greater end-to-end delay, our latency targets are rather meant to convey \textit{strict} and \textit{relaxed} latency requirements; that is, the precise latency target numbers may be scaled alongside a network transmission delay, and our track selection algorithm will operate in the same manner.

\subsection{Strict Latency Target}\label{sec:5ms_exp}

For our first evaluation, we targeted an end-to-end latency of 5 ms between event production and stream reconstruction at the receiver. Through sensitivity analysis, we found the optimal value for the number of events sent across each track for a 50 ms time window as $E=250$. Therefore, the maximum data rate of each track is 640 kbps. We found that higher values for $E$ make it impossible for the receiver to recover from spikes in latency, as the MoQ relay buffers become saturated. We set the relay bandwidth parameter $B \in \{1, 5, 25, 50, 100\}$ and varied the number of tracks, $N$, as  $N \in \{5, 10, 25\}$. Our results are shown in \cref{tab:adapt_results}.

\begin{figure}
    \centering
    \includegraphics[width=1\linewidth]{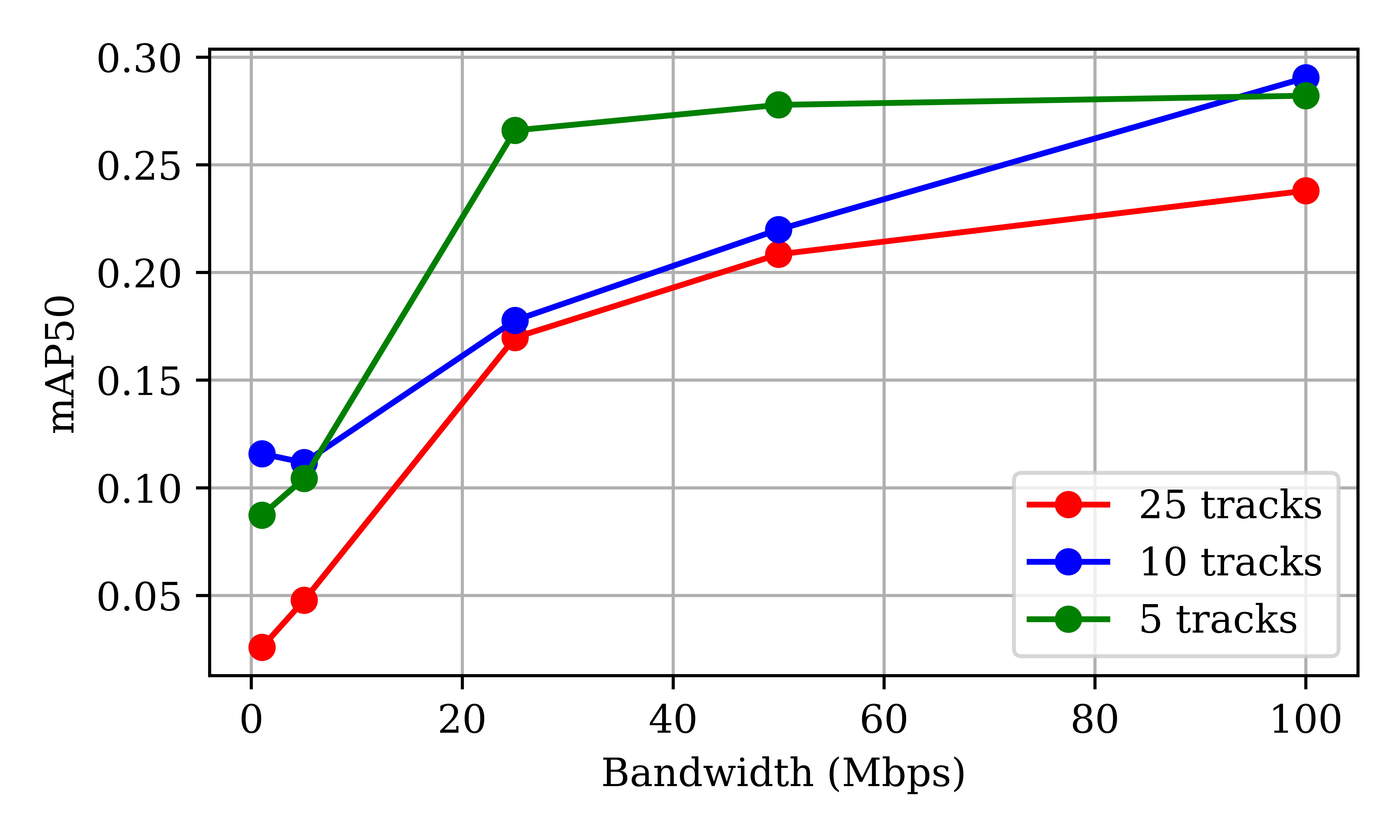}
    \caption{Representative example (from video \texttt{test\_night\_005}) of the strict latency experiments described in \cref{sec:5ms_exp}. As the network bandwidth decreases, the receiver has less data to pass to the RVT application, so object detection accuracy decreases. The configuration with 5 tracks performs best overall, avoiding the MoQ desynchronization issues that higher track counts incur. }
    \label{fig:strict_exp_bandwidth_map}
\end{figure}

We see that the configuration with $N=5$ yields the highest throughput for all bandwidth levels except $B=100$ Mbps. This result is counterintuitive, since one would expect that the throughput scales with $NE$. In practice, however, it is common for the various tracks to desynchronize at the MoQ relay. When the tracks desynchronize, the latency for stream reconstruction increases, as the receiver must wait to receive the data from the slowest subscribed track. Our algorithm then unsubscribes from one or more tracks, until the remaining tracks are synchronized and the latency is below the target threshold. \cref{fig:strict_exp_bandwidth_map} demonstrates this behavior, as the configuration with $N=5$ outperforms the others at nearly all bandwidth levels.

\cref{fig:time_example} illustrates the latency, event loss, and mAP metrics throughout the runtime of a particular video at $B=100$ Mbps. We see that our system can maintain a consistent latency measurement below the 5 ms target. The event loss plot (\cref{fig:time_example_strict_loss}) shows the combined effect of the source video event rate and the received data rate, based on the number of subscribed tracks. Alongside \cref{fig:time_example_strict_map}, we see that as the rate of event loss decreases, the mAP predictably increases. In particular, at our highest bandwidth setting, our mAP (averaged across all the videos) is reduced by as little as 0.36 compared to the baseline.

\begin{figure}
    \centering
    \begin{subfigure}{0.32\linewidth}
        \centering
        \includegraphics[width=\linewidth]{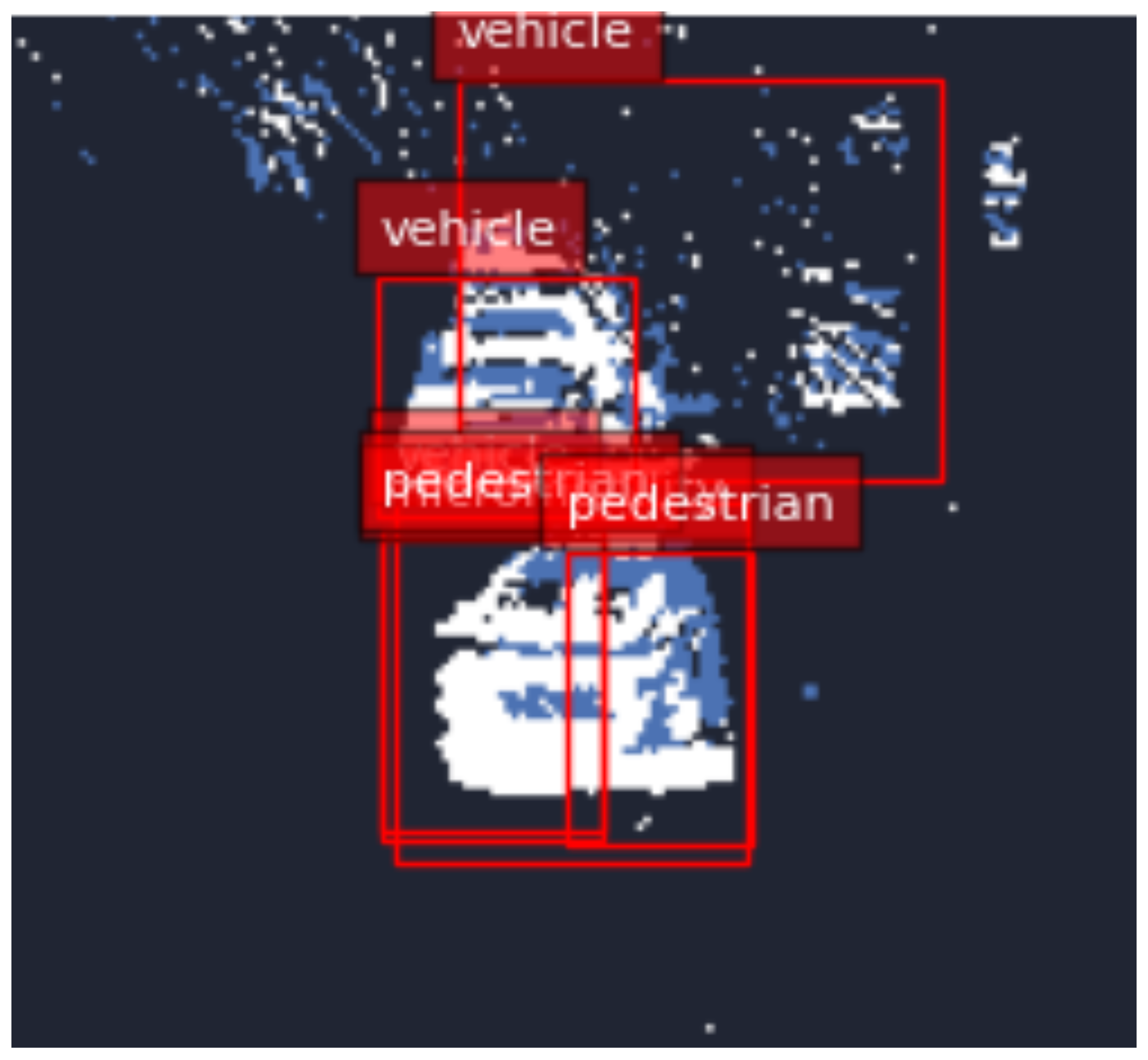}
        \caption{100 Mbps}
        \label{fig:zoom_100}
    \end{subfigure}
    \begin{subfigure}{0.32\linewidth}
        \centering
        \includegraphics[width=\linewidth]{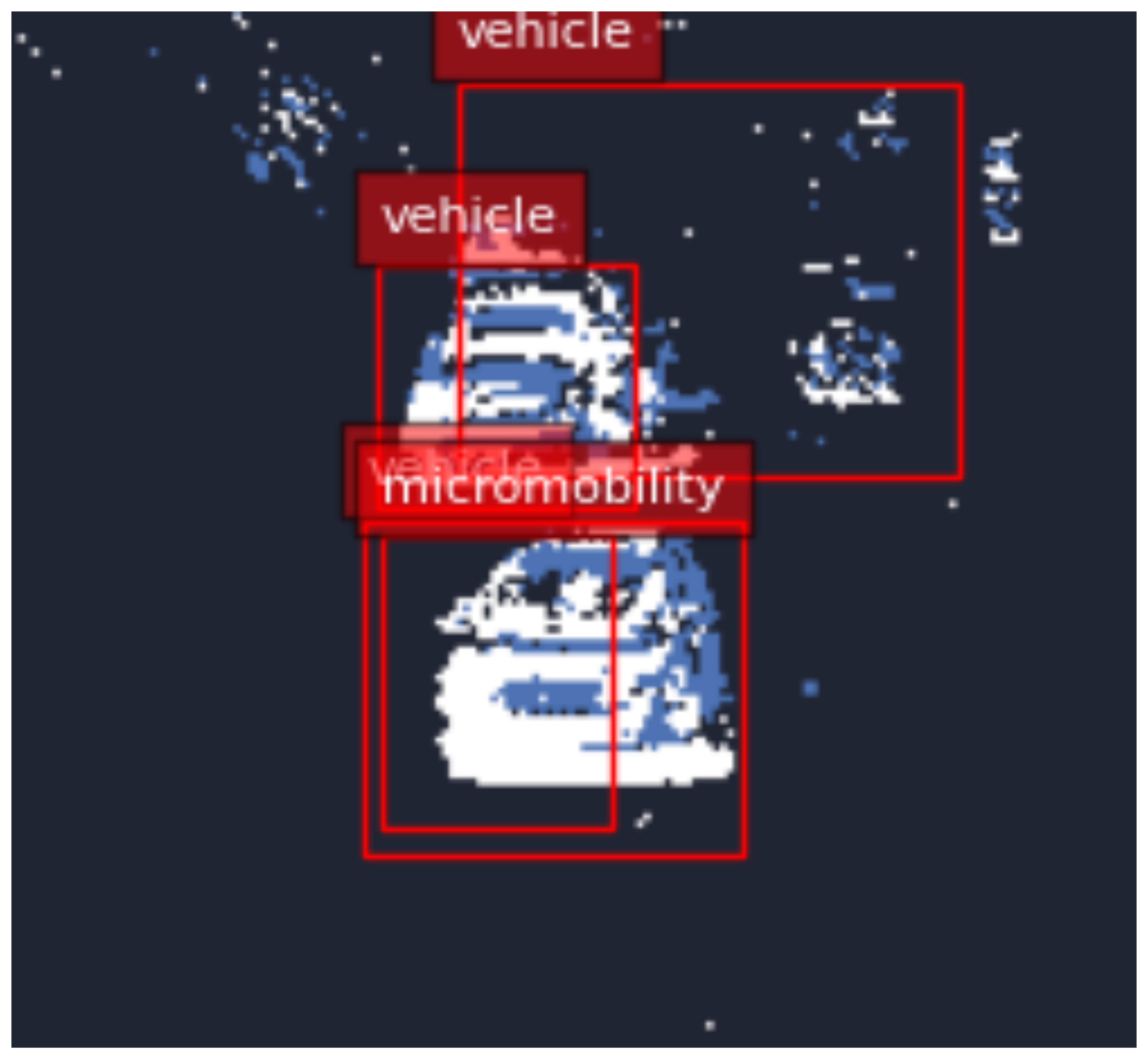}
        \caption{50 Mbps}
    \end{subfigure}
    \begin{subfigure}{0.32\linewidth}
        \centering
        \includegraphics[width=\linewidth]{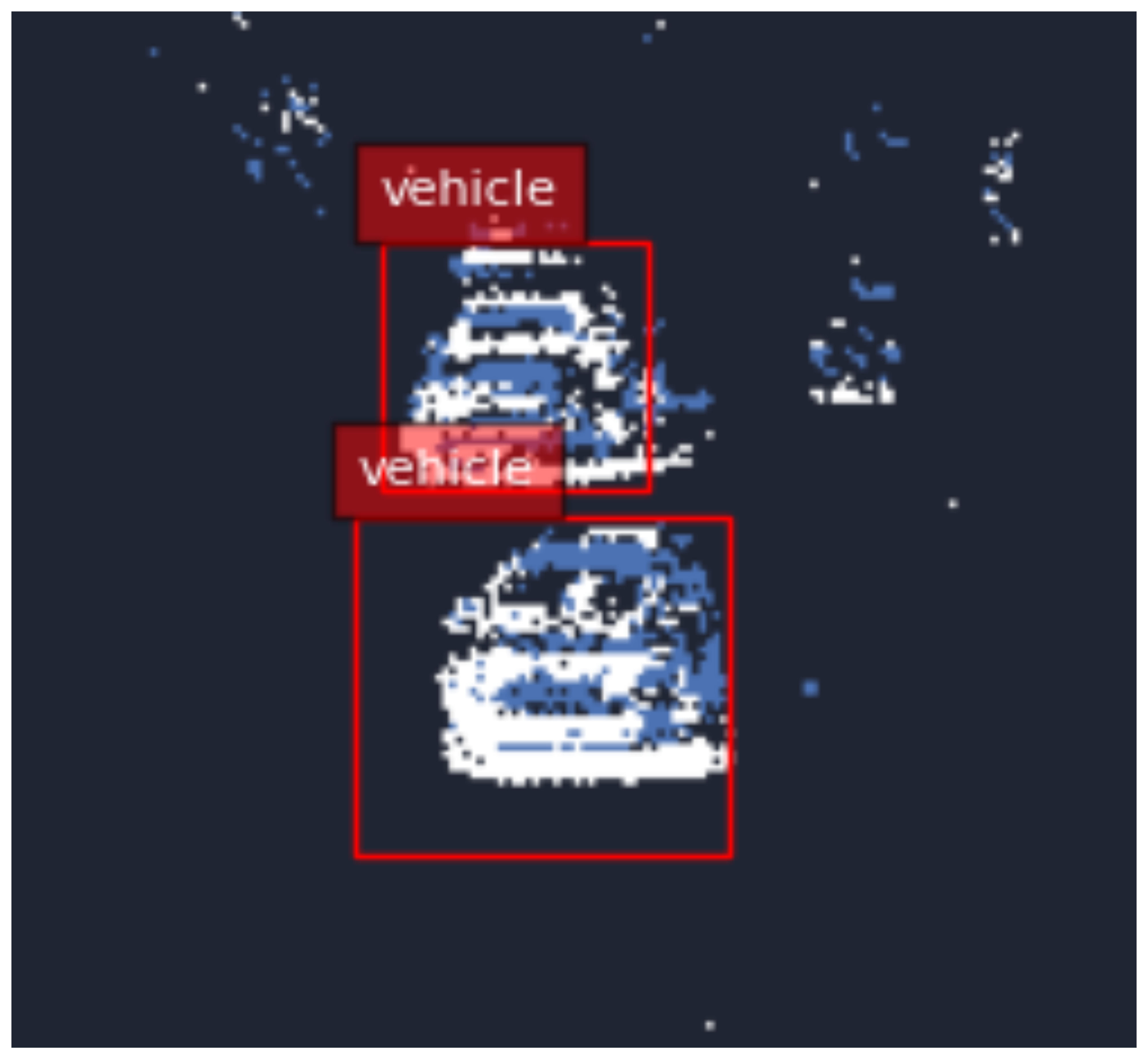}
        \caption{25 Mbps}
        \label{fig:zoom_25}
    \end{subfigure}
    \caption{Zoomed-in inference example from the fixed bitrate experiment of \cref{sec:fixed_bitrate}. In this, the indiscriminate loss of event data functions as a noise filter, \textit{improving} the accuracy of object detection.}
    \label{fig:zoom}
\end{figure}

\subsection{Relaxed Latency Target}\label{sec:50ms_exp}
Following our results for a strict latency requirement, we sought to quantify the effect of a higher latency target on our track selection algorithm. We selected a target of 50 ms, and found that our parameter for track size equally scaled to $E=2500$ events per track. At this rate, a single track at full capacity will transmit at a rate of 6.4 Mbps. As such, we cannot maintain a connection at bandwidth $B \in \{1,5\}$, and we removed those variations from our experiment configuration. We provide our results in \cref{tab:relaxed_results}.

We see that at $B=25$ our algorithm struggles to maintain the target latency, with a maximum latency of more than 8 seconds and a mean latency of $173.5$ ms. This implies that the track selection algorithm does not respond quickly enough to spikes in the source data rate (from the camera) under bandwidth-constrained settings. In contrast, at $B=50$ and $B=100$, the mean latency is comfortably below the 50 ms threshold, and our average mAP reduction is as low as 0.19 compared to the baseline. Even still, our maximum latency often exceeds the threshold. \cref{fig:time_example_source_rat} shows that in the later section of the example video, the source event rate increases dramatically. As in \cref{fig:time_example_relaxed_latency}, our adaptation scheme induces some thrashing in the latency: we unsubscribe from tracks to go below the target threshold, but then prematurely resubscribe. This phenomenon may be mitigated by adjusting the chunk size based on a moving average of the received data rate.

By visualizing the time-ordered data as in \cref{fig:time_example}, we further observed that our lossy streaming system occasionally outperforms the baseline in object detection accuracy. This result is especially surprising because this outperformance occurs even when the loss is greater than 75\%. In this case, our system seems to function as a \textit{noise filter}, removing extraneous events that impair RVT object detection. \cref{fig:zoom} provides an example, as both the bounding box accuracy and classification improves dramatically as the data rate decreases. However, we emphasize that greater event loss does not always correspond to an increase in mAP. In fact, the phenomenon is most prominent for the videos in our dataset with the highest overall data rate. The event loss then functions to normalize the event count histograms towards the training dataset. Qualitatively, we can see that the vehicles in \cref{fig:zoom_25} with a bandwidth of 25 Mbps appear very similar to \cref{fig:teaser_100} with a bandwidth of 100 Mbps. In contrast, the higher event density of \cref{fig:zoom_100} reflects a \textit{higher pixel sensitivity} of the event camera. Since the event pixel sensitivity is determined by a plethora of settings and environmental conditions \cite{delbruck_feedback_2021}, we see that there is room to explore rate-based normalization techniques at the application level. In addition to accommodating a diverse range of camera settings and capture conditions, such normalization may mitigate the consequences of streaming-driven data loss.

\section{Future Work}
As this work was merely the first foray into streaming systems for event camera data, there remains ample room for further research. Existing lossless event compression algorithms can be applied to each track, with a further evaluation of the difference in computational overhead of executing those processes in parallel rather than serially on a single event stream. The publisher should also have mechanisms to adapt its own sending rate according to the quality of the connection with the first relay. Furthermore, the realized throughput in our experiments is based on the fixed track send rate, $E$. To support a fully independent receiver, a mixture of SVC and ABR strategies may be implemented: the sender may publish complementary tracks of \textit{various} rates, to allow for both strict and relaxed latency requirements with arbitrary receivers. However, such a method will require a more robust track selection algorithm to accommodate the vastly different decode times for tracks of different sizes. Finally, the system should be evaluated with a live event camera on a real-world network, under subject categories other than the traffic monitoring scenario used in this work.

\section{Conclusion}

As event cameras continue to gain traction for computer vision research, we must bring to bear the systems-level concerns that are holding back widespread deployment. Towards this end, we proposed and evaluated a novel solution for receiver-driven adaptation of event camera streams. Across varied bandwidth conditions and latency requirements, our event stream partitioning and track selection algorithms unlock a tradeoff between latency and application performance. We demonstrated that we can meet a strict latency target of 5 ms, with object detection accuracy scaling along with the throughput. At a more relaxed latency target of 50 ms, we observe an average mAP reduction as low as 0.19 compared to the baseline system with no data loss. Importantly, we observed a varied effect of event loss on detection accuracy, as our system with high loss outperforms the baseline when the event pixel sensitivity is high, revealing an underlying weakness in the event vision application. While the techniques introduced in this work can be further refined, our research provides crucial insights for developing practical, distributed event camera systems for heterogeneous applications.

\bibliographystyle{ACM-Reference-Format}
\bibliography{references,references2}

\end{document}